\documentclass{llncs}
\usepackage[numbers, sort]{natbib}
\usepackage{amsmath, amssymb}
\usepackage{graphicx}
\usepackage[pagebackref=false,breaklinks=true,letterpaper=true,colorlinks,bookmarks=false]{hyperref}

\usepackage{xcolor}
\newcommand{\todo}[1]{}
\renewcommand{\todo}[1]{{\color{red} \textbf{{#1}}}}

\begin{document}

\title{Multi-modal Disease Classification in Incomplete Datasets Using Geometric Matrix Completion}
\titlerunning{Geometric matrix completion for multi-modal diagnosis}  
%
\author{Gerome Vivar\inst{1}\inst{2} \and Andreas Zwergal\inst{2} \and Nassir Navab\inst{1} \and Seyed-Ahmad Ahmadi\inst{2}}
%
\authorrunning{*}
%
%
\institute{Technical University of Munich (TUM), Munich, GER\\
\and 
German Center for Vertigo and Balance Disorders (DSGZ), Ludwig-Maximilians-Universit{\"a}t (LMU), Munich, GER
 }


\maketitle              

\begin{abstract}
In large population-based studies and in clinical routine, tasks like disease diagnosis and progression prediction are inherently based on a rich set of multi-modal data, including imaging and other sensor data, clinical scores, phenotypes, labels and demographics. However, missing features, rater bias and inaccurate measurements are typical ailments of real-life medical datasets. Recently, it has been shown that deep learning with graph convolution neural networks (GCN) can outperform traditional machine learning in disease classification, but missing features remain an open problem. In this work, we follow up on the idea of modeling multi-modal disease classification as a matrix completion problem, with simultaneous classification and non-linear imputation of features. Compared to methods before, we arrange subjects in a graph-structure and solve classification through geometric matrix completion, which simulates a heat diffusion process that is learned and solved with a recurrent neural network. We demonstrate the potential of this method on the ADNI-based TADPOLE dataset and on the task of predicting the transition from MCI to Alzheimer's disease. With an AUC of 0.950 and classification accuracy of 87\%, our approach outperforms standard linear and non-linear classifiers, as well as several state-of-the-art results in related literature, including a recently proposed GCN-based approach.
\end{abstract}
%
\section{Introduction}
\label{sec:intro}
In clinical practice and research, the analysis and diagnosis of complex phenotypes or disorders along with differentiation of their aetiologies rarely relies on a single clinical score or data modality, but instead requires input from various modalities and data sources. This is reflected in large datasets from well-known multi-centric population studies like the Alzheimer's Disease Neuroimaging Initiative (ADNI) and its derived TADPOLE grand challenge \footnote{\url{http://adni.loni.usc.edu} $\|$ \url{https://tadpole.grand-challenge.org/}}. TADPOLE data, for example, comprises demographics, neuropsychological scores, functional and morphological features derived from MRI, PET and DTI imaging, genetics, as well as histochemical analysis of cerebro-spinal fluid. The size and richness of such datasets makes human interpretation difficult, but it makes them highly suited for computer-aided diagnosis (CAD) approaches, which are often based on machine learning (ML) techniques \cite{ADNIMM:Zhang11,ADNIMM:Oishi11,ADNIMM:Moradi2015}. Challenging properties for machine learning include e.g. subjective, inaccurate or noisy measurements or a high number of features. Linear \cite{ADNIMM:Oishi11} and non-linear \cite{ADNIMM:Zhang11} classifiers for CAD show reasonable success in compensating for such inaccuracies, e.g. when predicting conversion from mild-cognitive-impairment (MCI) to Alzheimer's disease (AD). Recent work has further shown that an arrangement of patients in a graph structure based on demographic similarity \cite{GCN:Parisot17} can leverage network effects in the cohort and increase robustness and accuracy of the classification. This is especially valid when combined with novel methods from geometric deep learning \cite{GCN:Bronstein17}, in particular spectral graph convolutions \cite{GCN:Kipf16}. Similar to recent successes of deep learning methods in medical image analysis \cite{DLMIA:Litjens17}, deep learning on graphs shows promise for CAD, by modeling connectivity across subjects or features.\\
\indent Next to noise, a particular  problem of real-life, multi-modal clinical datasets is missing features, e.g. due to restrictions in examination cost, time or patient compliance. Most ML algorithms, including the above-mentioned, require feature-completeness, which is difficult to address in a principled manner \cite{IMP:Dong2013}. One interesting alternative to address missing features is to model CAD and disease classification as a matrix completion problem instead. Matrix completion was proposed in \cite{MC:Goldberg10} for simultaneously solving the three tasks of multi-label learning, transductive learning, and feature imputation. Recently, this concept was applied for CAD in multi-modal medical datasets for the first time \cite{SWMC:Thung16}, for prediction of MCI-to-AD conversion on ADNI data. The method introduced a pre-computed feature weighting term and outperformed linear classifiers on their dataset, however it did not yet leverage any graph-modeled network effects of the population as in \cite{GCN:Parisot17}. 
To this end, several recent works incorporated a geometric graph structure into the matrix completion problem \cite{GMC:Kalofolias14,RGCN:Monti17, MC:Rao15}. All these methods were applied on non-medical datasets, e.g. for recommender systems \cite{RGCN:Monti17}. Hence, their goal was solely imputation, without classification. Here, we unify previous ideas in a single stream-lined method that can be trained end-to-end.\\
\indent \textbf{Contribution.}
In this work, we follow up on the idea of modeling multi-modal CAD as a matrix completion problem \cite{MC:Goldberg10} with simultaneous imputation and classification \cite{SWMC:Thung16}. We leverage cohort network effects by integrating a population graph with a solution based on geometric deep learning and recurrent neural networks \cite{RGCN:Monti17}.
For the first time, we demonstrate geometric matrix completion (GMC) and disease classification from multi-modal medical data, towards MCI-to-AD prediction from TADPOLE features at baseline examination. In this difficult task, GMC significantly outperforms regular linear and non-linear machine learning methods as well as three state-of-the-art results from related works, including a recent approach based on graph-convolutional neural networks.\\



%
\section{Methods}
\label{sec:methods}
%

%
\subsection{Dataset and Preprocessing}
As an example application, we utilize the ADNI-based TADPOLE dataset, with the goal of predicting whether an MCI subject will convert to AD given their baseline information. We select all unique subjects with baseline measurements from ADNI1, ADNIGO, and ADNI2 in the TADPOLE dataset which were diagnosed as MCI including those diagnosed as EMCI and LMCI. Following \cite{SWMC:Thung16}, we retrospectively label those subjects whose condition progressed to AD within 48 months as cMCI and those whose condition remained stable as sMCI. The remaining MCI subjects who progressed to AD after month 48 are excluded from this study. We use multi-modal features from MRI, PET, DTI, and CSF at baseline, i.e. excluding longitudinal features. We use all numerical features from this dataset to stack with the labels and include age and gender to build the graph, following the intuition and methodology from \cite{GCN:Parisot17}.  

\subsection{Matrix Completion}
We will start by describing the matrix completion problem. Suppose there exists a matrix $\mathbf{Y} \in \mathbb{R}^{m\times n}$ where the values in this matrix are not all known. The goal is to recover the missing values in this matrix. A well-defined description of this problem is to assume that the matrix is of low rank \cite{MC:Candes08},
\begin{equation}
\label{eq:lrmc}
	\min_{\mathbf{X} \in \mathbb{R}^{m \times n}}\ \operatorname{rank}(\mathbf{X})\ \operatorname{s.t.}\  x_{ij} = y_{ij}, \forall ij \in \Omega,
\end{equation}

where $\mathbf{X}$ is the $m \times n$ matrix with values $x_{ij}$, $\Omega$ is the set of known entries in matrix $\mathbf{Y}$ with $y_{ij}$ values. However, this rank minimization problem \eqref{eq:lrmc} is known to be computationally intractable. So instead of solving for rank($\mathbf{X}$), we can replace it with its convex surrogate known as the nuclear norm $||\mathbf{X}||_*$ which is equal to the sum of its singular values \cite{MC:Candes08}. In addition, if the observations in Y have noise, the equality constraint in equation \eqref{eq:lrmc} can be replaced with the squared Frobenius norm $||.||_{\operatorname{F}}^2$ \cite{MC:Candes10},

\begin{equation}
\label{eq:lrmcfb}
	\min_{\mathbf{X} \in \mathbb{R}^{m\times n}}\ ||\mathbf{X}||_*\ 
    + \ \frac{\gamma}{2} ||\mathbf{\Omega} \circ (\mathbf{Y} - \mathbf{X})||_{\operatorname{F}}^2 \ ,
\end{equation}

where $\mathbf{\Omega}$ is the masking matrix of known entries in $\mathbf{Y}$ and $\circ$ is the Hadamard product. Alternatively, a factorized solution to the representation of the matrix $\mathbf{X}$ was also introduced in \cite{MC:Srebro05,MC:Rao15}, as the formulation using the full matrix makes it hard to scale up to large matrices such as the famous Netflix challenge. Here, the matrix $\mathbf{X} \in \mathbb{R}^{m\times n}$ is factorized into 2 matrices $\mathbf{W}$ and $\mathbf{H}$ via SVD, where $\mathbf{W}$ is $m \times r$ and $\mathbf{H}$ is $n \times r$, with $r \ll \operatorname{min}(m,n)$. \citet{MC:Srebro05} showed that the nuclear norm minimization problem can then be rewritten as: 

\begin{equation}
\label{eq:mc_factorized}
\min_{W, H} \frac{1}{2}||\mathbf{W}||_{\operatorname{F}}^2 + \frac{1}{2}||\mathbf{H}||_{\operatorname{F}}^2 + \frac{\gamma}{2}||\mathbf{\Omega} \circ (\mathbf{W}\mathbf{H}^T - \mathbf{X})||_{\operatorname{F}}^2
\end{equation}

\begin{figure}[t]
  \centering
  \includegraphics[width=1.0\textwidth]{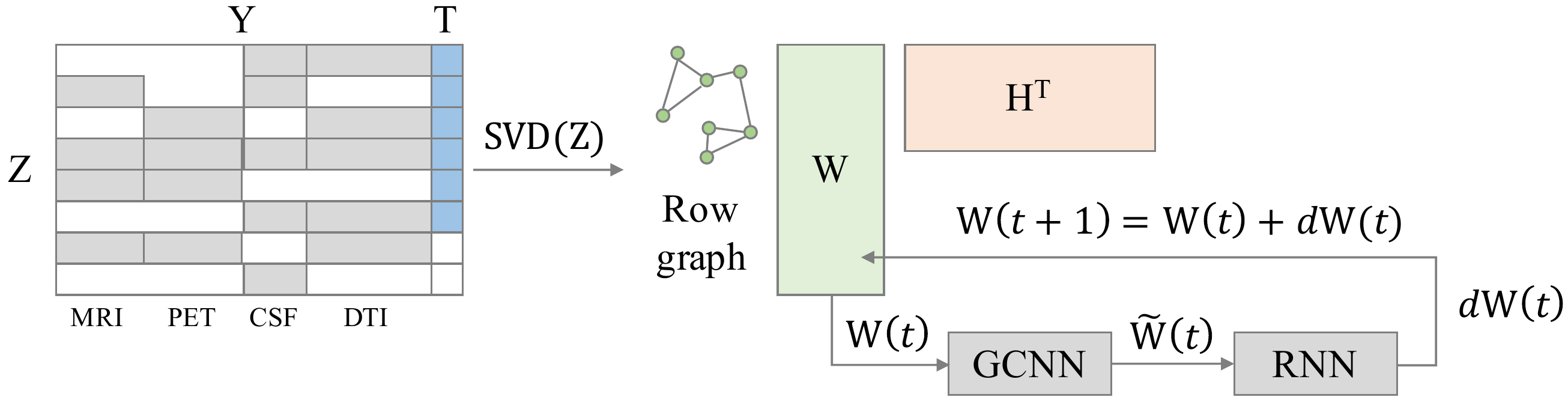}\\
  \caption{Illustration of the overall approach: the matrix $Z$ comprising incomplete features and labels is factorized into $Z=WH^T$. A connectivity graph is defined over rows W. During optimization, GCNN filters are learned along with RNN parameters and weight updates for $W$, towards optimal matrix completion of $Z$ and simultaneous inference of missing features and labels in the dataset.}
  \label{fig:Model}
\end{figure}

\subsection{Matrix Completion on Graphs}
The previous matrix completion problem can be extended to graphs \cite{MC:Rao15,GMC:Kalofolias14}. Given a matrix $\mathbf{Y}$, we can assume that the rows/columns of this matrix are on the vertices of the graph \cite{GMC:Kalofolias14}. This additional information can then be included into the matrix completion formulation in equation \eqref{eq:lrmcfb} as a regularization term \cite{GMC:Kalofolias14}. To construct the graph, we can use meta-information out of these rows/columns or use the row/column vectors of this matrix to calculate a similarity metric between pairs of vertices. Given that every row in the matrix has this meta-information, \citet{GMC:Kalofolias14} showed that we can build an undirected weighted row graph $G_r = (V_r, E_r, A_r)$, with vertices $V_r = \{1,...,m\}$. Edges $E_r \subseteq V_r \times V_r$ are weighted with non-negative weights represented by an adjacency matrix $A_r \in \mathbb{R}^{m\times m}$. The column graph $G_c = (V_c, E_c, A_c)$ is built the same way as the row graph, where the columns are now the vertices in $G_c$. \citet{GMC:Kalofolias14} showed that the solution to this problem is equivalent to adding the Dirichlet norms, $||\mathbf{X}||_{D,r}^2 = \operatorname{tr}(X^T L_r X)$ and $||\mathbf{X}||_{D,r}^2 = \operatorname{tr}(X L_c X^T)$, where $L_r$ and $L_c$ are the unnormalized row and column graph Laplacian, to equation \eqref{eq:lrmcfb},

\begin{equation}
\label{eq:gmc}
	\min_{\mathbf{X} \in \mathbb{R}^{m \times n}}\ ||\mathbf{X}||_*\ 
    +\ \frac{\gamma}{2} ||\mathbf{\Omega} \circ (\mathbf{Y} - \mathbf{X})||_{\operatorname{F}}^2 \ 
    +\ \frac{\alpha_r}{2}||\mathbf{X}||_{D,r}^2 +\ \frac{\alpha_c}{2}||\mathbf{X}||_{D,c}^2 \
\end{equation}
The factorized formulation \cite{MC:Rao15,RGCN:Monti17} of equation \eqref{eq:gmc} is
\begin{equation}
\label{eq:gmc_factorized}
\min_{\mathbf{W}, \mathbf{H}} \frac{1}{2}||\mathbf{W}||_{\operatorname{D,r}}^2 + \frac{1}{2}||\mathbf{H}||_{\operatorname{D,c}}^2 + \frac{\gamma}{2}||\mathbf{\Omega} \circ (\mathbf{Y} - \mathbf{W}\mathbf{H}^T)||_{\operatorname{F}}^2
\end{equation}

%
\subsection{Geometric Matrix Completion with Separable Recurrent Graph Neural Networks}
%

In \cite{RGCN:Monti17}, Monti et al. propose to solve the matrix completion problem as a learnable diffusion process using Graph Convolutional Neural Networks (GCNN) and Recurrent Neural Networks (RNN). The main idea here is to use GCNN to extract features from the matrix and then use LSTMs to learn the diffusion process. They argue that combining these two methods allows the network to predict accurate small changes $\mathbf{dX}$ (or $\mathbf{dW}$, $\mathbf{dH}$ of the matrices $\mathbf{W}$, $\mathbf{H}$) to the matrix $\mathbf{X}$. Further details regarding the main ideas in geometric deep learning have been summarized in a review paper \cite{GCN:Bronstein17}, where they elaborate how to extend convolutional neural networks to graphs. Following \cite{RGCN:Monti17}, we use Chebyshev polynomial basis on the factorized form of the matrix $\mathbf{X} = \mathbf{W}\mathbf{H}^{\operatorname{T}}$ to represent the filters on the respective graph to each matrix $\mathbf{W}$ and $\mathbf{H}$. In this work, we only apply GCNN to the matrix $\mathbf{W}$ as we only have a row graph and leave the matrix $\mathbf{H}$ as a changeable variable. Figure \ref{fig:Model} illustrates the overall approach.

\subsection{Geometric Matrix Completion for Heterogeneous Matrix Entries}
In this work, we propose to solve multi-modal disease classification as a geometric matrix completion problem. We use a Separable Recurrent GCNN (sRGCNN) \cite{RGCN:Monti17} to simultaneously predict the disease and impute missing features on a dataset which has partially observed features and labels. Following \citet{MC:Goldberg10}, we stack a feature matrix $\mathbf{Y} \in \mathbb{R}^{m\times n}$ and a label matrix $\mathbf{T} \in \mathbb{R}^{m\times c}$ as a matrix $\mathbf{Z} \in \mathbb{R}^{m \times n+c}$, where m is the number of subjects, n is the dimension of the feature matrix, and c is the dimension of the target values. In the TADPOLE dataset, we stack the $m \times n$ feature matrix to the $m \times 1$ label matrix, where the feature matrix contains all the numerical features and the label matrix contains the encoded binary class labels for cMCI and sMCI. We build the graph by using meta-information from the patients such as their age and gender, similar to  \cite{GCN:Parisot17}, as these information are known to be risk factors for AD. We compare two row graph construction approaches, first from age and gender information using a similarity metric \cite{GCN:Parisot17} and second from age information only, using Euclidian distance-based k-nearest neighbors. Every node in a graph corresponds to a row in the matrix $\mathbf{W}$, and the row values to its associated feature vector. Since we only have a row graph, we leave the matrix $\mathbf{H}$ to be updated during backpropagation. To run the geometric matrix completion method we use the loss:

\begin{equation}
\label{eq:gcn_seperable}
\ell(\Theta) = \frac{\gamma_a}{2}||\mathbf{W}||_{\operatorname{D,r}}^2 + \frac{\gamma_b}{2}||\mathbf{W}||_{\operatorname{F}}^2 + \frac{\gamma_c}{2}||\mathbf{H}||_{\operatorname{F}}^2 + \frac{\gamma_d}{2}||\mathbf{\Omega}_{a} \circ (\mathbf{Z} - \mathbf{W}\mathbf{H}^T)||_{\operatorname{F}}^2 + \gamma_e(\ell_{\mathbf{\Omega}_{b}}(\mathbf{Z}, \mathbf{X})),
\end{equation}

where $\Theta$ are the learnable parameters, where $\mathbf{Z}$ denotes the target matrix, $\mathbf{X}$ is the approximated matrix, $||.||_{D,r}^2$ denotes the Dirichlet norm on a normalized row graph Laplacian, $\mathbf{\Omega}_{a}$ denotes the masking on numerical features, $\mathbf{\Omega}_{b}$ is the masking on the classification labels, and $\ell$ is the binary cross-entropy.

%
\section{Results}
\label{sec:res}

\begin{figure}[b!]
  \centering
  \includegraphics[width=0.8\textwidth]{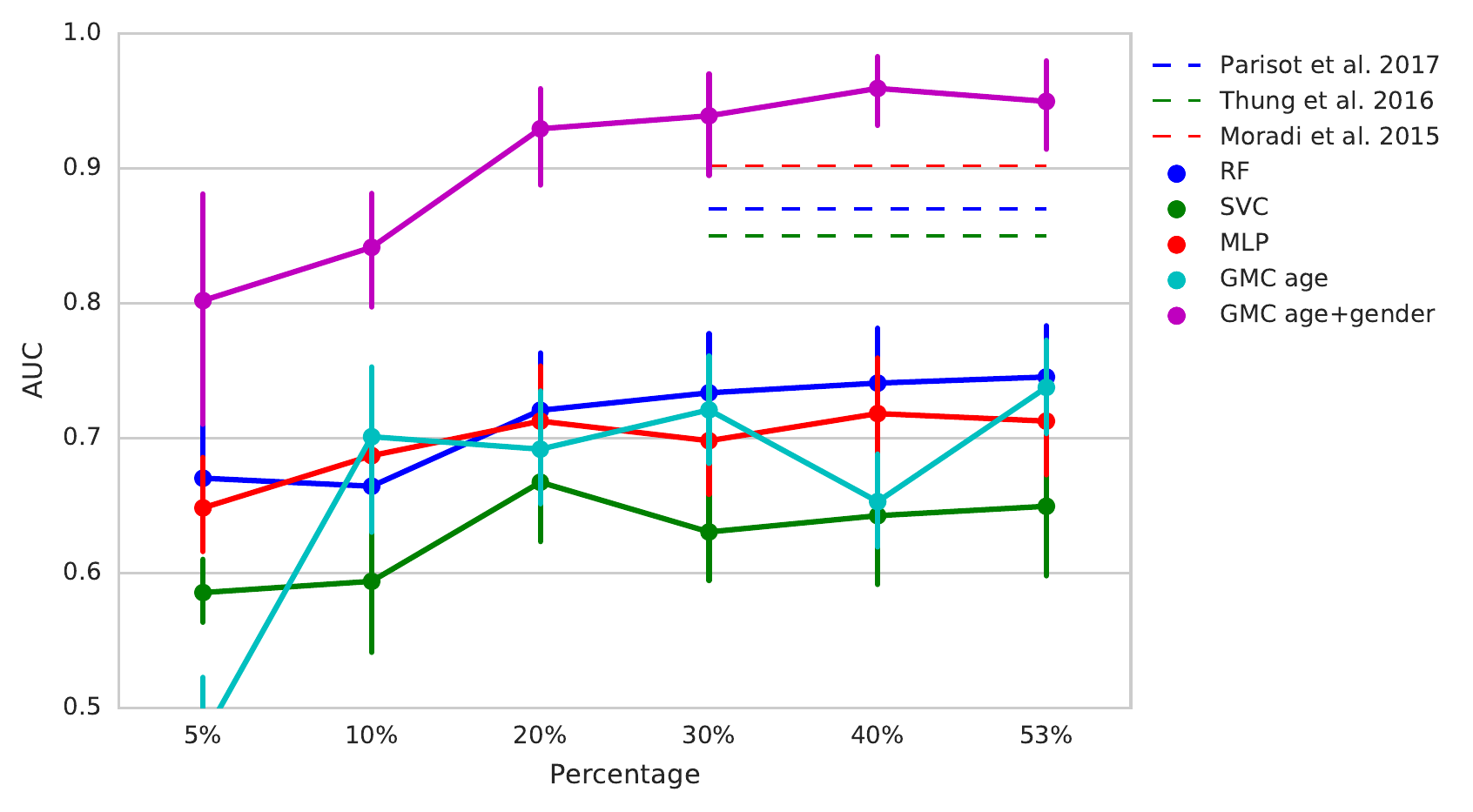}\\
  \caption{Classification results: Area under the curve (AUC) of our method, for different amounts of feature-completeness and in comparison to linear/non-linear standard methods, and three state-of-the-art results in literature (\citet{GCN:Parisot17}, \citet{SWMC:Thung16}, \citet{ADNIMM:Moradi2015}).}
  \label{fig:AUC}
\end{figure}
We evaluate our approach on multi-modal TADPOLE data (MRI, PET, CSF, DTI) to predict MCI-to-AD conversion  and compare it to several other multi-modal methods as baseline. We use a stratified 10-fold cross-validation strategy for all methods. Hyperparameters were optimized using Hyperopt \footnote{\url{http://hyperopt.github.io/hyperopt/}}, through nested cross-validation, targeting classification loss (binary cross-entropy) on a hold-out validation set (10\% in each fold of training data). Following \cite{RGCN:Monti17}, we use the same sRGCNN architecture with parameters: rank=156, chebyshev polynomial order=18, learning rate=0.00089, hidden-units=36, $\gamma_a$=563.39, $\gamma_b$=248.91, $\gamma_c$=688.85, $\gamma_d$=97.63, and $\gamma_e$=890.14.

It is noteworthy that at baseline, the data matrix $Y$ with above-mentioned features is already feature-incomplete, i.e. only 53\% filled. We additionally reduce the amount of available data randomly to 40\%, 30\% etc. to 5\%. Figure \ref{fig:AUC} shows a comprehensive summary of our classification results in terms of  area-under-the-curve (AUC). Methods we compare include mean imputation with random forest (RF), linear SVM (SVC) and multi-layer-perceptron (MLP), as well as three reference methods from literature \cite{ADNIMM:Moradi2015, GCN:Parisot17, SWMC:Thung16}, which operated on slightly different selections of ADNI subjects and on all available multi-modal features. While implementations of \cite{SWMC:Thung16, ADNIMM:Moradi2015} are not publicly available, we tried to re-evaluate the method  \cite{GCN:Parisot17} using their published code. Unfortunately, despite our best efforts and hyperparameter optimization on our selection of TADPOLE data, we were not able to reproduce any AUC value close to their published value. To avoid any mistake on our side, we provide the reported AUC results rather than the worse results from our own experiments. 

At baseline, our best-performing method with a graph setup based on age and gender ("GMC age-gender") \cite{GCN:Parisot17} achieves classification with an AUC value of 0.950, compared to 0.902 \cite{ADNIMM:Moradi2015}, $\sim0.87 $ \cite{GCN:Parisot17} and 0.851 \cite{SWMC:Thung16}. In terms of classification accuracy,  we achieved a value of 87\%, compared to 82\% \cite{ADNIMM:Moradi2015} and 77\% \cite{GCN:Parisot17} (accuracy not reported in \cite{SWMC:Thung16}). Furthermore, our method significantly outperforms standard classifiers RF, MLP and SVC  at all levels of matrix completeness. The second graph configuration for our method ("GMC age" only) performs significantly worse and less stable than ("GMC age-gender"), confirming the usefulness of the row graph construction based on the subject-to-subject similarity measure proposed in \cite{GCN:Parisot17}. Due to lower complexity of the GMC approach \cite{RGCN:Monti17}, training a single fold on recent hardware (Tensorflow on Nvidia GTX 1080 Ti) is on average 2x faster (11.8s) than GCN (25.9s) \cite{GCN:Parisot17}.
%
\section{Discussion and Conclusion}
\label{sec:dis}
%
In this paper, we proposed to view disease classification in multi-modal but incomplete clinical datasets as a geometric matrix completion problem. As an exemplary dataset and classification problem, we chose MCI-to-AD prediction. Our initial results using this method show that GMC outperforms three competitive results from recent literature in terms of AUC and accuracy. At all levels of additional random dropout of features, GMC also outperforms standard imputation and classifiers (linear and non-linear). There are several limitations which are worthy to be addressed. Results in Figure \ref{fig:AUC} demonstrate that GMC is still sensitive to increasing amounts of feature incompleteness, in particular at feature presence below 15\%. This may be due to our primary objective of disease classification during hyper-parameter optimization. For the same reason, we did not evaluate the actual imputation performed by GMC. However, an evaluation in terms of RMSE and a comparison to principled imputation methods \cite{IMP:Dong2013} would be highly interesting, if this loss is somehow incorporated during hyperparameter optimization. Furthermore, we only evaluated GMC on ADNI data as represented in the TADPOLE challenge, due to the availability of multiple reference AUC/accuracy values in literature. As mentioned, however, disease classification in high-dimensional but incomplete datasets with multiple modalities is an abundant problem in computer-aided medical diagnosis. In this light, we believe that the promising results obtained through GMC in this study are of high interest to the community. 
\\

\paragraph{Acknowledgments.} The study was supported by the German Federal Ministry of Education and Health (BMBF) in connection with the foundation of the German Center for Vertigo and Balance Disorders (DSGZ) (grant number 01 EO 0901).

%

{
\bibliographystyle{plainnat}
\bibliography{Bibliography.bib}

\begin{thebibliography}{16}
\providecommand{\natexlab}[1]{#1}
\providecommand{\url}[1]{\texttt{#1}}
\expandafter\ifx\csname urlstyle\endcsname\relax
  \providecommand{\doi}[1]{doi: #1}\else
  \providecommand{\doi}{doi: \begingroup \urlstyle{rm}\Url}\fi

\bibitem[Bronstein et~al.(2017)Bronstein, Bruna, Lecun, Szlam, and
  Vandergheynst]{GCN:Bronstein17}
Michael~M. Bronstein, Joan Bruna, Yann Lecun, Arthur Szlam, and Pierre
  Vandergheynst.
\newblock {Geometric Deep Learning: Going beyond Euclidean data}.
\newblock \emph{IEEE Signal Processing Magazine}, 34\penalty0 (4):\penalty0
  18--42, 2017.

\bibitem[Candes and Recht(2008)]{MC:Candes08}
E.J. Candes and B.~Recht.
\newblock {Exact low-rank matrix completion via convex optimization}.
\newblock \emph{46th Annual Allerton Conference on Communication, Control, and
  Computing}, pages 1--49, 2008.

\bibitem[Candes and Plan(2010)]{MC:Candes10}
Emmanuel~J. Candes and Yaniv Plan.
\newblock {Matrix completion with noise}.
\newblock \emph{Proceedings of the IEEE}, 98\penalty0 (6):\penalty0 925--936,
  2010.

\bibitem[Dong and Peng(2013)]{IMP:Dong2013}
Y.~Dong and C.~Y. Peng.
\newblock {{P}rincipled missing data methods for researchers}.
\newblock \emph{Springerplus}, 2\penalty0 (1):\penalty0 222, Dec 2013.

\bibitem[Goldberg et~al.(2010)Goldberg, Recht, Xu, Nowak, and
  Zhu]{MC:Goldberg10}
Andrew Goldberg, Ben Recht, Junming Xu, Robert Nowak, and Xiaojin Zhu.
\newblock Transduction with matrix completion: Three birds with one stone.
\newblock In \emph{Advances in Neural Information Processing Systems (NIPS)},
  pages 757--765. 2010.

\bibitem[Kalofolias et~al.(2014)Kalofolias, Bresson, Bronstein, and
  Vandergheynst]{GMC:Kalofolias14}
Vassilis Kalofolias, Xavier Bresson, Michael Bronstein, and Pierre
  Vandergheynst.
\newblock Matrix completion on graphs.
\newblock \emph{arXiv:1408.1717}, 2014.

\bibitem[Kipf and Welling(2016)]{GCN:Kipf16}
Thomas~N. Kipf and Max Welling.
\newblock Semi-supervised classification with graph convolutional networks.
\newblock \emph{CoRR}, arXiv:1609.02907, 2016.

\bibitem[Litjens et~al.(2017)Litjens, Kooi, Bejnordi, Setio, Ciompi,
  Ghafoorian, van~der Laak, van Ginneken, and Sánchez]{DLMIA:Litjens17}
Geert Litjens, Thijs Kooi, Babak~Ehteshami Bejnordi, Arnaud Arindra~Adiyoso
  Setio, Francesco Ciompi, Mohsen Ghafoorian, Jeroen~A.W.M. van~der Laak, Bram
  van Ginneken, and Clara~I. Sánchez.
\newblock A survey on deep learning in medical image analysis.
\newblock \emph{Medical Image Analysis}, 42:\penalty0 60 -- 88, 2017.

\bibitem[Monti et~al.(2017)Monti, Bronstein, and Bresson]{RGCN:Monti17}
Federico Monti, Michael~M. Bronstein, and Xavier Bresson.
\newblock Geometric matrix completion with recurrent multi-graph neural
  networks.
\newblock \emph{CoRR}, arXiv:1704.06803, 2017.

\bibitem[Moradi et~al.(2015)Moradi, Pepe, Gaser, Huttunen, and
  Tohka]{ADNIMM:Moradi2015}
Elaheh Moradi, Antonietta Pepe, Christian Gaser, Heikki Huttunen, and Jussi
  Tohka.
\newblock {Machine learning framework for early MRI-based Alzheimer's
  conversion prediction in MCI subjects}.
\newblock \emph{NeuroImage}, 104:\penalty0 398--412, 2015.

\bibitem[Oishi et~al.(2011)Oishi, Akhter, Mielke, Ceritoglu, Zhang, Jiang, Li,
  Younes, Miller, van Zijl, Albert, Lyketsos, and Mori]{ADNIMM:Oishi11}
Kenichi Oishi, Kazi Akhter, Michelle Mielke, Can Ceritoglu, Jiangyang Zhang,
  Hangyi Jiang, Xin Li, Laurent Younes, Michael Miller, Peter van Zijl, Marilyn
  Albert, Constantine Lyketsos, and Susumu Mori.
\newblock {Multi-Modal MRI Analysis with Disease-Specific Spatial Filtering:
  Initial Testing to Predict Mild Cognitive Impairment Patients Who Convert to
  Alzheimer’s Disease}.
\newblock \emph{Frontiers in Neurology}, 2:\penalty0 54, 2011.

\bibitem[Parisot et~al.(2017)Parisot, Ktena, Ferrante, Lee, Moreno, Glocker,
  and Rueckert]{GCN:Parisot17}
Sarah Parisot, Sofia~Ira Ktena, Enzo Ferrante, Matthew Lee, Ricardo~Guerrerro
  Moreno, Ben Glocker, and Daniel Rueckert.
\newblock Spectral graph convolutions for population-based disease prediction.
\newblock In \emph{International Conference on Medical Image Computing and
  Computer-Assisted Intervention}, pages 177--185, 2017.

\bibitem[Rao et~al.(2015)Rao, Yu, Ravikumar, and Dhillon]{MC:Rao15}
Nikhil Rao, Hsiang-Fu Yu, Pradeep Ravikumar, and Inderjit~S Dhillon.
\newblock {Collaborative Filtering with Graph Information: Consistency and
  Scalable Methods}.
\newblock \emph{Neural Information Processing Systems (NIPS)}, pages 1--9,
  2015.

\bibitem[Srebro et~al.(2005)Srebro, Rennie, and Jaakkola]{MC:Srebro05}
Nathan Srebro, Jason D~M Rennie, and Tommi~S Jaakkola.
\newblock {Maximum-Margin Matrix Factorization}.
\newblock \emph{Advances in Neural Information Processing Systems (NIPS)},
  17:\penalty0 1329--1336, 2005.

\bibitem[Thung et~al.(2016)Thung, Adeli, Yap, and Shen]{SWMC:Thung16}
Kim-Han Thung, Ehsan Adeli, Pew-Thian Yap, and Dinggang Shen.
\newblock Stability-weighted matrix completion of incomplete multi-modal data
  for disease diagnosis.
\newblock In \emph{International Conference on Medical Image Computing and
  Computer-Assisted Intervention}, pages 88--96, 2016.

\bibitem[Zhang et~al.(2011)Zhang, Wang, Zhou, Yuan, and Shen]{ADNIMM:Zhang11}
Daoqiang Zhang, Yaping Wang, Luping Zhou, Hong Yuan, and Dinggang Shen.
\newblock Multimodal classification of alzheimer's disease and mild cognitive
  impairment.
\newblock \emph{NeuroImage}, 55\penalty0 (3):\penalty0 856 -- 867, 2011.

\end{thebibliography}
}

\end{document}